\documentclass[11pt,a4paper]{article}
\usepackage[hyperref]{acl2021}
\usepackage{times}
\usepackage{latexsym}

\usepackage{microtype}

\usepackage{booktabs} 
\usepackage{graphicx} 
\usepackage{colortbl} 
\usepackage{multirow} 
\usepackage{bbding} 
\usepackage{makecell} 
\usepackage{bm} 
\usepackage{color} 
\usepackage{xcolor} 
\usepackage{textcomp,booktabs} 
\usepackage{amsmath} 
\usepackage{amsfonts} 
\usepackage{amsthm,amsmath,amssymb} 
\usepackage{mathrsfs} 
\usepackage{subfigure}
\usepackage{mathtools}
\definecolor{mygray}{gray}{.9}
\aclfinalcopy 
\def\aclpaperid{389} 
\title{Retrieve \& Memorize: Dialog Policy Learning with Multi-Action Memory} 
\author{Yunhao Li\textsuperscript{1}, Yunyi Yang\textsuperscript{1}, Xiaojun Quan\textsuperscript{1}\thanks{\;\;Corresponding author}, Jianxing Yu\textsuperscript{2}\\
  \textsuperscript{1}School of Computer Science and Engineering, Sun Yat-sen University, China \\
  \textsuperscript{2}School of Artificial Intelligence, Sun Yat-sen University, China \\
  \texttt{\{liyh355,yangyy37\}@mail2.sysu.edu.cn, \{quanxj3,yujx26\}@mail.sysu.edu.cn} \\}

\date{}
\begin{document}
\maketitle
\begin{abstract}
\label{sec:abstract}
Dialogue policy learning, a subtask that determines the content of system response generation and then the degree of task completion, is essential for task-oriented dialogue systems. However, the unbalanced distribution of system actions in dialogue datasets often causes difficulty in learning to generate desired actions and responses. In this paper, we propose a retrieve-and-memorize framework to enhance the learning of system actions. Specially, we first design a neural context-aware retrieval module to retrieve multiple candidate system actions from the training set given a dialogue context. Then, we propose a memory-augmented multi-decoder network to generate the system actions conditioned on the candidate actions, which allows the network to adaptively select key information in the candidate actions and ignore noises. We conduct experiments on the large-scale multi-domain task-oriented dialogue dataset MultiWOZ 2.0 and  MultiWOZ 2.1.~Experimental results show that our method achieves competitive performance among several state-of-the-art models in the context-to-response generation task.
\end{abstract}

\section{Introduction}
\label{sec:introduction}

Task-oriented dialogue systems communicate with users through natural language conversations to accomplish a wide range of tasks such as restaurant and flight bookings.
Recent years have seen a rapid growth of interest in building task-oriented dialogue systems \cite{Budzianowski2018}.  Such systems are usually decomposed into several subtasks, including natural
language understanding \cite{gupta2018semantic}, dialogue state tracking \cite{zhong-etal-2018-global},
\begin{figure}[h]
    \centering
    \includegraphics[width=0.48\textwidth]{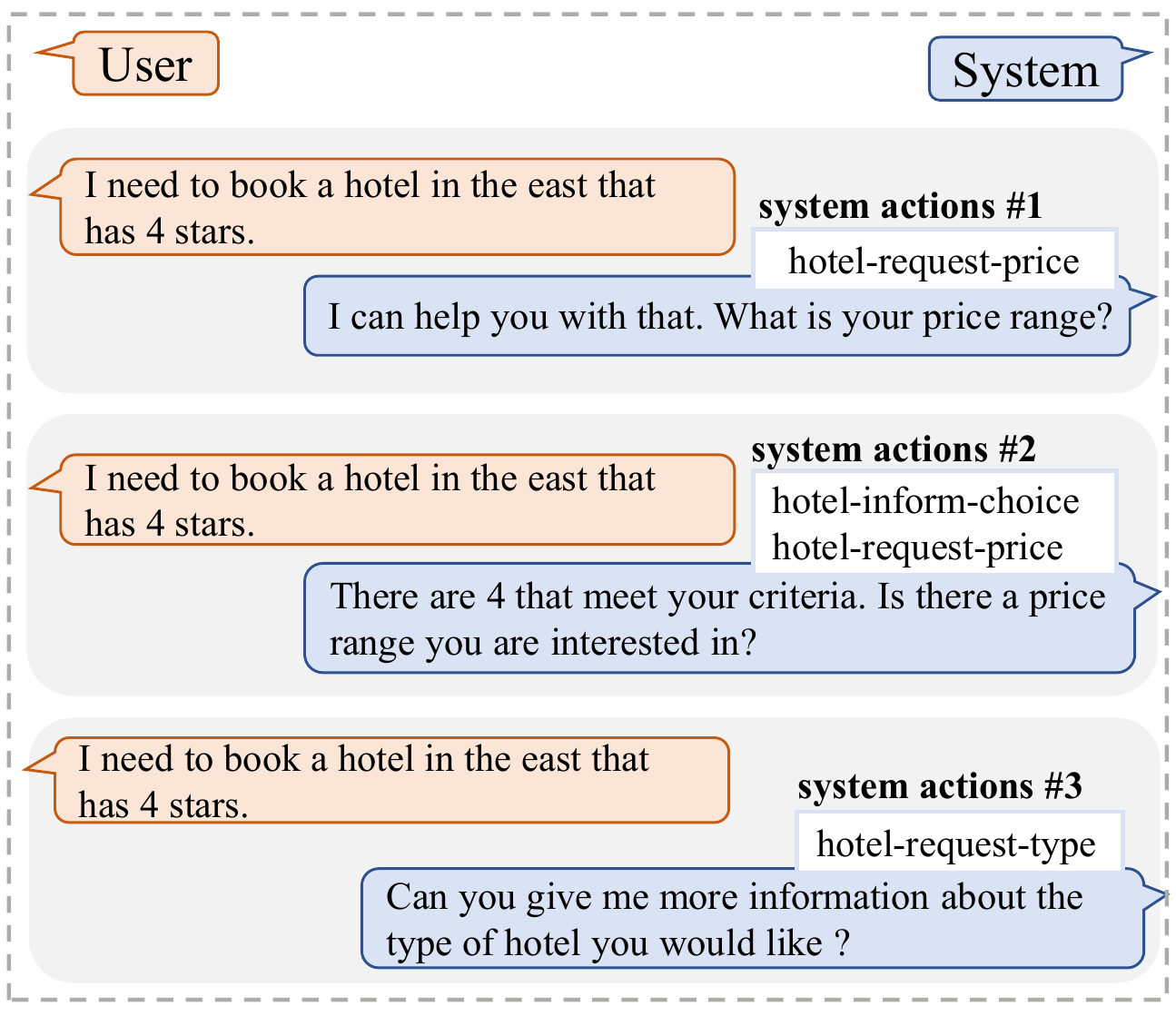}
    \caption{An example of the one-to-many property,where there are multiple appropriate system actions and responses given the same dialogue context.}
    \label{fig:example}
\end{figure}
system actions (dialogue policy) prediction, and response generation \cite{wen2015semantically,chen2019semantically,Zhao2019}, where system actions can be viewed as a semantic plan of response generation. 
One of the main challenges for context-to-response generation in task-oriented dialogue systems comes from the intrinsic one-to-many property in conversations. As shown in Figure \ref{fig:example}, there can be multiple valid system actions for the same dialogue context, which means that multiple satisfactory system responses can be generated correspondingly. However, in most collected dialogue datasets, each dialogue context has only one reference, which leads to an unbalanced distribution of system actions and responses in multi-domain dialogue datasets \cite{zhang2020task}. Models trained on such unbalanced datasets tend to overfit high-frequency system actions and underfit low-frequency ones. 

One line of work focuses on the representation of system actions, which alleviates the unbalanced problem to a certain extent.
\citet{chen2019semantically} reconstruct system actions into a compact graph representation. \citet{Zhao2019} treat system actions as latent variables and use reinforcement learning to optimize them. \citet{wang-etal-2020-multi-domain} model system actions prediction as a sequence generation problem by treating system actions as a sequence of tokens.
On the other hand, \citet{zhang2020task} explicitly modeling the one-to-many property to enrich system action diversity through a rule-based multi-action data augmentation. Specifically, they treat system actions that follow the same dialogue state as alternative valid actions and train them together with the reference system action.
However, their data augmentation framework has two shortcomings.
First, it enforces a rigid mapping between dialogue state and system actions. Dialogue state, which consists of information such as belief state and user actions, is not flexible enough to represent the whole dialogue context and thus limits the diversity of the mapped system actions.
Second, they treat the mapped system actions as gold references during training which may force the model to fit noise in the mapped system actions and ultimately hinder the quality of the generated system actions.

To address the above limitations, we propose to model the one-to-many property more effectively by retrieving multiple candidate system actions and selectively taking the candidates into consideration when generating system action.
We design a retrieve-and-memorize framework that consists of a context-aware neural retrieval module (CARM) and a memory-augmented multi-decoder network (MAMD).
Specifically, the context-aware retrieval module uses a pre-trained language model to convert the dialogue history as well as belief state into a context representation of each sample. 
Multiple candidate system actions are retrieved based on the distances between the context vector and the representations of other samples in the latent space. 
These retrieved candidate actions are more diverse and consistent with the dialogue context since they are obtained based on a more holistic representation.
Instead of treating the candidates impartially with the gold references, we encode them into a memory bank and the memory-augmented multi-decoder network can dynamically attend to the memory bank during system actions generation.
Additionally, we employ a random sampling mechanism where during training, the memory bank is filled with randomly sampled system actions with a probability, which allows the model to learn to distinguish the quality of the candidates and adaptively adjust its dependence on the candidate actions.

We evaluate our model on MultiWOZ \cite{Budzianowski2018}, a large-scale multi-domain dataset for task-oriented dialogue systems. Extensive experiments and analyses are conducted to demonstrate the effectiveness of our model, and the results show that it significantly outperforms the baseline model. Our main contributions are summarized as follows:
\begin{itemize}
\item We propose a context-aware retrieval module that can retrieve multiple appropriate system actions given a dialogue context.
\item We propose a memory-augmented multi-decoder network that can generate system actions based on multiple candidate actions.
\item Our model outperforms several state-of-the-art baselines on a large-scale multi-domain dataset for task-oriented dialogue systems.
\end{itemize}

\section{Related Work}
\label{sec:related work}

One line of research focuses on the representation of system actions. A typical approach to encoding system actions is by concatenating the one-hot representation at each level of actions into a flat vector \cite{wen2015semantically,Budzianowski2018}. Such sparse representations make the learning of system actions difficult. To overcome the sparsity issue, \citet{chen2019semantically} compact the one-hot vector representation based on the intrinsic hierarchical structures of system actions, and apply hierarchical disentangled self-attention to generate system response. \citet{Zhao2019} treat system actions as latent variables and use reinforced learning \cite{he2016deep} to optimize them. 
Recently, \citet{wang-etal-2020-multi-domain} 
propose a co-generation framework to generate system actions and response sequentially, which achieves a new state of the art in the context-to-response task.
Our proposed framework adopts the idea of modeling belief state and system actions \cite{wang-etal-2020-multi-domain,liang2020moss} as sequences and generates the belief state, system action, and response sequentially to make better use of the intermediate supervision. 

Another line of research uses data augmentation to expand the training data. \citet{gao-etal-2020-paraphrase} use the paraphrase technique \cite{li2019decomposable,wang2019task} to generate user utterances and then expand the training set with the augmented user utterances. 
\citet{zhang2020task} augment system actions with a mapped dialogue state, which consists of belief state, user action, turn domain, and database search result.
Such mapping is rule-based and requires user actions for the construction of dialogue state, which takes extra annotations.
Both of the above approaches treat the augmented samples as equivalent to the gold ones, which may force the model to fit noises in the augmented data. 
In this paper, we focus on a better neural retrieval method for the alternative system actions, and instead of directly training on the augmented actions, we encode them in a memory bank as auxiliary information.

\section{Methodology}
\label{sec:method}
To frame the problem of dialogue policy learning, we use $X_t=\{U_1,..,U_{t-1},R_{t-1},U_t\}$ to denote the dialogue history at turn $t$ of a multi-turn conversation, where $U_i = u_1u_2,...u_{m_i}$ and $R_i = r_1r_2...r_{n_i}$ are the $i$-th user utterance and system response, respectively. Following previous works \cite{zhang2020task,liang2020moss}, we convert the belief state and system actions from a list of triples to sequences. 
For example, the belief state ``\textit{restaurant-food-Chinese,restaurant-price-expansive}" is converted to ``\textit{restaurant [food] Chinese [price] expansive}", and the system actions ``\textit{restaurant-inform-price,restaurant-inform-phone}" are converted to ``\textit{restaurant [inform] price phone}". We use $B_t=b_1b_2...b_p$ and $A_t=a_1a_2...a_q$ to represent the current belief state and system action, respectively. Our goal is to generate system actions $A_t$ and system response $R_t$ of turn $t$ based on the dialogue context $X_t$ and belief state $B_t$.

We employ a retrieve-and-memorize framework to generate the system response.~First, we use a context-aware retrieve module to retrieve multiple proper candidate system actions from the training set. Then, we encode the candidate actions into a memory bank and propose a memory-augmented module to enhance the action generation.
\begin{figure}[ht]
    \centering
    \includegraphics[width=0.46\textwidth]{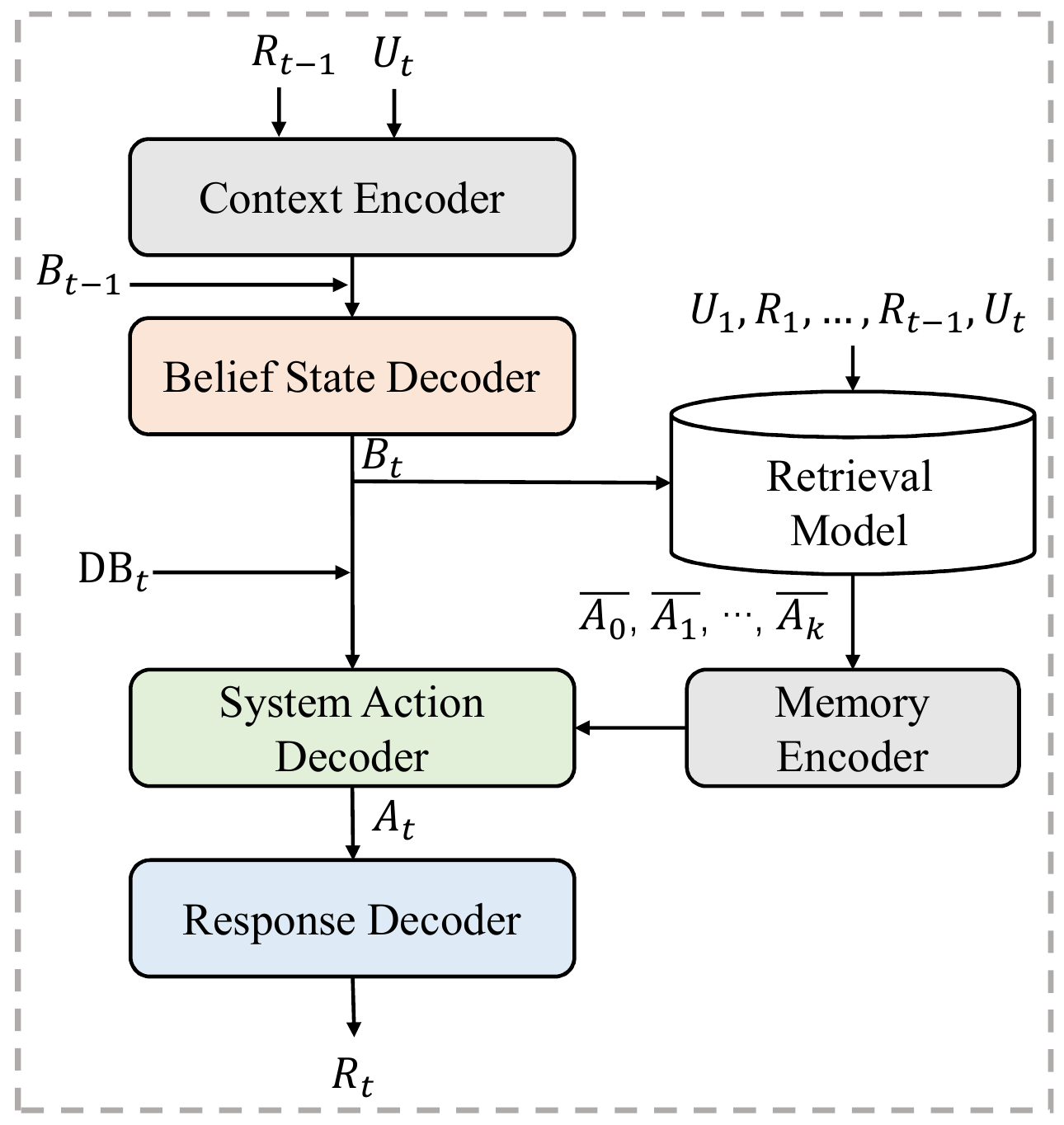}
     \caption{An overview of the proposed model.}
\label{fig:model_over_view}
\end{figure}
\subsection{Context-Aware Retrieval Module}
In order to retrieve alternative system actions that are more comprehensive and context-aware, we utilize the powerful pre-trained language model BERT \cite{devlin2019bert} to obtain distributed representations of the dialogue context.
We search in the training corpus for system actions with similar distributed representations and retrieve them as alternative candidate actions. 
Concretely, we combine the dialogue history $X_t=\{U_1,..,U_{t-1},R_{t-1},U_t\}$ and belief state $B_t$ as dialogue context and feed the concatenated dialogue context into a pre-trained BERT encoder:
\begin{equation}
\begin{aligned}
H = & \text{BERT}([CLS]\oplus B_t \oplus [SEP] \oplus X_t ), \\
\end{aligned}
\end{equation}
where $\oplus$ is the concatenation operator, $[CLS]$ is a special token that precedes every input sequence of BERT, and $[SEP]$ is a special token used to separate different parts of the input sequence. 
The BERT model encodes the input dialogue context into a sequence of hidden states $H=\{h^{CLS},h^{1},...,h^{L}\}$. 
We use $h^{CLS}$ to represent the distributed representation of dialogue context, since $h^{CLS}$ is expected to capture the information of the whole sequence. 
Then we use $L_2$ distance to measure the similarity between the distributed representations of different dialog contexts: 
\vspace{-0.3cm}
\begin{equation}
    L_2(h^{CLS}_{i},h^{CLS}_{j}) = ||h^{CLS}_{i}-h^{CLS}_{j}||_2.
\end{equation}
Based on the $L_2$ distance, $k$ most similar dialogue contexts are selected from the training set, and the corresponding system actions constitute a candidate actions set $ \{\bar{A_1},\bar{A_2},...,\bar{A_k}\}$.

\noindent \textbf{Pre-training Task}
Directly applying $h^{CLS}$ from BERT without fine-tuning or further pre-training may not result in desired dialogue context representations that correlate well with system actions.
A good dialogue contextual representation should satisfy the property that dialogue contexts with similar semantics are close to each other in the representation space. 
Therefore, we further pre-train the BERT model with an actions prediction task:
\begin{equation}
    \begin{aligned}
    p(y|B_t,X_t) = \text{classifier}(h^{CLS}),
    \end{aligned}
\end{equation}
where $y \in \mathbb{R}^{D}$ is a one-hot label of system actions \cite{chen2019semantically}, $D$ is the dimension of the label space, and \textit{classifier} is a simple linear classifier.

\subsection{Memory-Augmented Multi-Decoder Network}
We propose a memory-augmented multi-decoder network that jointly generates belief state, system actions, and system response while having access to a memory bank when generating the system action.
Given the retrieved candidate system actions, we encode these candidates into the memory bank and enhance the generation of system actions by querying the memory bank during decoding.

\noindent \textbf{Encoding Module} 
We use Bidirectional GRUs \cite{chung2014empirical} as our encoders. First, we encode the current user utterance, the previous system response and the previous belief state separately into hidden states:
\begin{equation}
    \begin{aligned}
        &H_u = \text{Encoder}(U_t),\\
        &H_{pre\_r} = \text{Encoder}(R_{t-1}),\\
        &H_{pre\_b} = \text{Encoder}(B_{t-1}),
    \end{aligned}
\end{equation}
Then, another encoder is used to encode the candidate system actions into memory bank:
\begin{equation}
    M_t = \text{Encoder}_M(\bar{A_1} \oplus \bar{A_2} \oplus... \oplus \bar{A_k}), \\
\end{equation}
where $M_t=\{m_1,...,m_k\}$.

\noindent \textbf{Belief State Generation} The belief state $B_t$ of turn \textit{t} is generated based on the current user utterance $U_t$, previous system response $R_{t-1}$ and previous belief state $B_{t-1}$. 
The generation of $B_t$ at each time step $\tau$ can be formulated as follows:
\begin{equation}
    \begin{aligned}
        &s_\tau = \text{Attn}(h_{\tau-1},H_u,H_{pre\_r},H_{pre\_b}), \\
        &c_\tau = [s_\tau \oplus e(b_{\tau-1})], \\
            &p(b_{\tau}|b_{1:\tau-1}), h_\tau =\text{Dec}_b(c_\tau,h_{\tau-1},H_{pre\_b}), \\
    \end{aligned}
\end{equation}
where $Attn$\footnote{Please refer to the appendix for more details.\label{formula}} is an attention function, $e(b_{\tau-1})$ is the embedding of the previous token, $h_{\tau-1}$ is the hidden state from the last decoding step, and $h_0 = \textbf{0}$. $Dec_b$\textsuperscript{\ref {formula}} is the belief state decoder augmented with copy mechanism \cite{gu2016incorporating}, which can copy tokens from the previous belief state. $p(b_{\tau}|b_{1:\tau-1})$ is a 
distribution over vocabulary. We use cross entropy between ground truth and the output distribution $\mathcal{L}_b(\theta)$ as the loss of belief state generation. We collect the hidden states $H_b=\{h_0,h_1,...,h_p\}$ of each step to feed them into the action decoder.

\begin{figure}[t]
        \centering
        \includegraphics[width=0.46\textwidth]{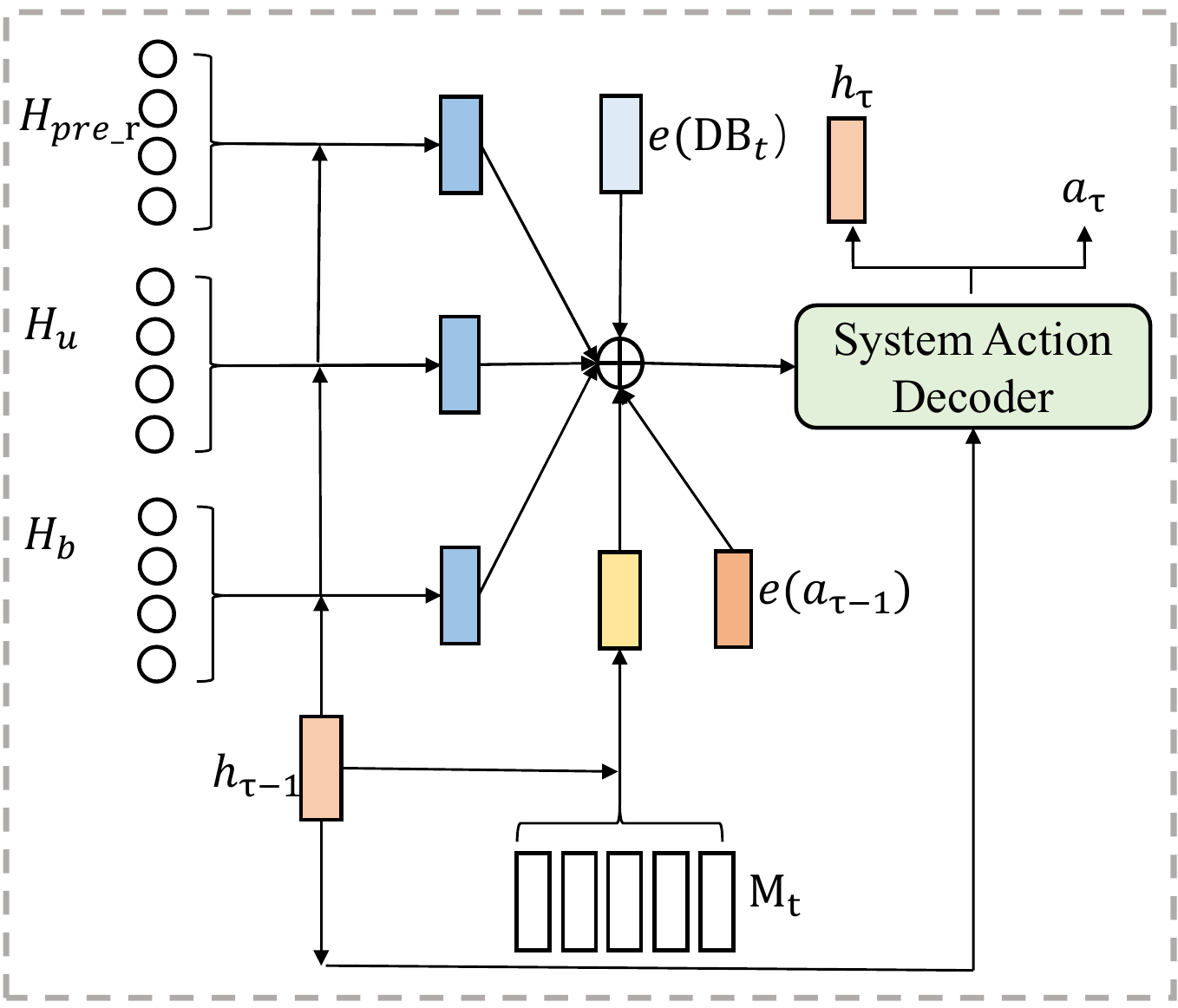}
\caption{Generation of system actions at time step $\tau$.}
\label{fig:decodeing}
\end{figure}

\noindent \textbf{Memory-Augmented Action Generation} 
As shown in Figure \ref{fig:decodeing}, the system action $A_t$ of turn \textit{t} is generated based on not only the dialog history and the current belief state, but also the memory bank which encodes the retrieved candidate system actions.
For the generation of $A_t$, at each time step, we first compute the state $s_\tau$:
\begin{equation}
        s_\tau = \text{Attn}(h_{\tau-1},H_u,H_{pre\_r},H_b). \\
\end{equation}
Then, we use the hidden state $h_{\tau-1}$ to query the encoded candidate system actions memory $M_t$:
\begin{equation}
    \begin{aligned}
        &a_{\tau}^i = \text{tanh}(W[h_{\tau-1} \oplus m_i]),\\
        &{\alpha}_{\tau} = \text{Softmax}(a_{\tau}),\\
        &v_\tau = \sum\nolimits_{i=1}^{k}{{\alpha}_{\tau}^i m_i},\\
    \end{aligned}
\end{equation}
where $W$ are learnable parameters and $v_\tau$ contains information from the memory. Now we incorporate $v_\tau$ into the generation process:
\begin{equation}
    \begin{aligned}
    &c_\tau = [s_\tau \oplus e(a_{\tau-1}) \oplus e(DB_t) \oplus v_{\tau}],\\
    &p(a_{\tau}|a_{1:\tau-1}), h_\tau = \text{Dec}_a(c_\tau,h_{\tau-1},H_b),\\
    \end{aligned}
\end{equation}
where $e(a_{\tau-1})$ is the embedding of the previous token, $e(DB_t)$ is the embedding of the database search result which indicates the number of matched entities. $Dec_a$ is the action decoder augmented with copy mechanism. The cross entropy $\mathcal{L}_a(\theta)$ between the output distribution and ground truth is the loss of actions generation. We collect the hidden states $H_a=\{h_0,h_1,...,h_q\}$ as well to feed it into the system response decoder. 

\noindent\textbf{Random Sampling}
Though the retrieved candidate system actions are considered to be of high quality and suitable given the dialogue context, we would still like our model to avoid taking those candidates for granted and developing excessive dependence on them.
To this end, during training, the memory bank is filled with randomly sampled system actions with a probability $p$, and retrieved candidates with a probability $(1-p)$.
This allows the model to learn to distinguish good candidates from bad candidates.

\noindent\textbf{Response Generation} Lastly, we generate the system response conditioned on the hidden states of user utterance $H_u$, belief state $H_b$ and system actions $H_a$ with the response decoder $Dec_r$:
\begin{equation}
    \begin{aligned}
        &s_\tau = \text{Attn}(h_{\tau-1},H_u,H_b,H_a), \\
        &c_\tau = [s_\tau \oplus e(r_{\tau-1})], \\
        &p(r_{\tau}|r_{1:\tau-1}), h_\tau = \text{Dec}_r(c_\tau,h_{\tau-1},H_b), \\
    \end{aligned}
\end{equation}
The response generation loss $\mathcal{L}_r(\theta)$ is the cross entropy between the output and ground truth.

\noindent\textbf{Objective Function} The final objective function is the sum of belief state loss, actions generation loss and response generation loss:
\begin{equation}
    \mathcal{L}(\theta) = \mathcal{L}_b(\theta) + \mathcal{L}_a(\theta) + \mathcal{L}_r(\theta)
\end{equation}

\section{Experiments}
\label{sec:experimental setting}

\subsection{Dataset and Metrics}

We conduct our experiments primarily on MultiWOZ 2.0 \cite{Budzianowski2018}.
It consists of 8438 dialogues spanning several domains and topics. 
Each of the test and validation sets contains 1000 dialogues.
As for automatic evaluation, we use \textit{Inform Rate} and \textit{Success Rate} to evaluate dialogue task completion. The former measures whether the system has provided a proper entity and the latter measures whether it has answered all the requested attributes \cite{Budzianowski2018}.~Besides, \textit{BLEU} \cite{papineni2002bleu} is used to measure the fluency of generated responses.~To measure the overall quality, we compute a combined score by $(\textit{Inform}~\hspace{-0.13cm}+\hspace{-0.13cm}~\textit{Success})\hspace{-0.1cm}\times\hspace{-0.15cm}~0.5~\hspace{-0.13cm}+\hspace{-0.13cm}~\textit{BLEU}$ \cite{mehri2019structured}.

\begin{table*}[h]
\small
    \centering
    \begin{tabular}{l p{0.7cm}<{\centering} p{0.7cm}<{\centering} p{1.1cm}<{\centering} p{1.1cm}<{\centering} p{1.1cm}<{\centering} c}
    \toprule
        \textbf{Model} & \textbf{DA} & \textbf{LM} & \textbf{Inform} & \textbf{Success} & \textbf{BLEU} & \textbf{Combined Score} \\
    \midrule
        SC-LSTM \cite{wen2015semantically}  & \XSolidBrush & \XSolidBrush & 74.50 & 62.50 & 20.50 & 89.00 \\
    
        LaRL \cite{Zhao2019}& \XSolidBrush & \XSolidBrush & 82.80 & 79.20 & 12.80 & 94.10 \\
   
        SimpleTOD \cite{hosseini2020simple} & \XSolidBrush & \Checkmark & 88.90 & 67.10 & 16.90 & 94.90\\    
        
        HDSA \cite{chen2019semantically} & \XSolidBrush & \Checkmark & 82.90 & 68.90 & \textbf{23.60} & 99.50 \\
        
        DAMD \cite{zhang2020task} & \XSolidBrush & \XSolidBrush & 89.50 & 75.80 & 18.30 & 100.90 \\
        
        DAMD (aug) \cite{zhang2020task} & \Checkmark & \XSolidBrush & 89.20 & 77.90 & 18.60 & 102.15 \\
        
        PARG \cite{gao-etal-2020-paraphrase}& \Checkmark & \XSolidBrush & 91.10 & 78.90 & 18.80 & 103.80 \\
        
        
        MarCo \cite{wang-etal-2020-multi-domain} & \XSolidBrush & \Checkmark & 92.30 & 78.60 & 20.02 & 105.47 \\
        
        UBAR \cite{yang2020ubar} & \XSolidBrush & \Checkmark & 94.00 & 83.60 & 17.20 & 106.00 \\
        
        LAVA \cite{lubis-etal-2020-lava} & \XSolidBrush & \XSolidBrush & \textbf{97.50}&	\textbf{94.80}&	12.10&	108.25 \\
        
        HDNO \cite{wang2020modelling} & \XSolidBrush & \XSolidBrush & 96.40&	84.70&	18.85&	109.37 \\
       \rowcolor{mygray} 
       MAMD  & \Checkmark & \XSolidBrush & 95.70 & 88.90 & 18.90 & \textbf{111.20} \\
    \bottomrule
    \end{tabular}
    \caption{Overall results on the MultiWOZ 2.0 dataset. \emph{DA} indicates whether to use data augmentation, and \emph{LM} indicates whether to use pre-trained language models to predict system action.}
    \label{tab:overall_results}
    \vspace{-0.3cm}
\end{table*}

\begin{table}[h]
\small
\centering
\setlength{\tabcolsep}{2.0mm}{
    \begin{tabular}{l  c c c c}
    \toprule
        \textbf{Model} & \textbf{Inform} & \textbf{Success} & \textbf{BLEU} & \textbf{Score} \\
    \midrule
        SimpleTOD & 85.10 & 73.50 & 16.22 & 95.52 \\ 
        
        HDSA & 86.30 & 70.60 & \textbf{22.36} & 100.81 \\
        
        
        MarCo & 92.50 & 77.80 & 19.54 & 104.69 \\
        UBAR   & 92.70&	81.00&	16.70&	103.55 \\
        
        LAVA   & \textbf{96.39}&	83.57&	14.02&	104.00 \\
        
        HDNO   & 92.80&	83.00&	18.97&	106.87 \\
        \rowcolor{mygray} 
        MAMD & 94.20 & \textbf{86.20} & 18.80 & \textbf{109.00} \\
    \bottomrule
    \end{tabular}
}
    \caption{Overall results on the MultiWOZ 2.1 dataset.}
    \label{tab:overall_results_2.1}
\end{table}

\subsection{Implementation Details}

Our model is trained on a 12 GB Nvidia GeForce RTX 2080 Ti with a batch size of 80. Our implementation\footnote{https://github.com/yunhaoli1995/MAMD-TOD} is based on PyTorch \cite{NEURIPS2019_bdbca288}. 
We pre-trained the BERT model based on the open-source library Transformers \cite{wolf-etal-2020-transformers}.
The dimension of word embeddings is 50 and the hidden size is 100.~We use one-layer Bidirectional GRUs \cite{chung2014empirical} as context encoders and three GRUs augmented with copy mechanism as decoders. The candidate actions are encoded by another  Bidirectional GRU. 
We use Adam \cite{kingma2015adam} optimizer with a learning rate of 0.007. 
We use greedy search to decode system actions and beam search with a beam size of 5 to decode system responses.
We use the ground truth belief states for a fair comparison with other baselines. 
We train our model for 60 epochs and select the best model on the validation set, and then evaluate it on the test set to get the final results.

\subsection{Baselines}
We compare our full model MAMD with several baselines on MultiWOZ 2.0:
SC-LSTM \cite{wen2015semantically}, LaRL \cite{Zhao2019}, HDSA \cite{chen2019semantically} , DAMD \cite{zhang2020task}, PARG \cite{gao-etal-2020-paraphrase}, SimpleTOD \cite{hosseini2020simple}, MarCo \cite{wang-etal-2020-multi-domain}, UBAR \cite{yang2020ubar}, HDNO \cite{wang2020modelling}, LAVA \cite{lubis-etal-2020-lava}.
Especially, SC-LSTM and HDSA treat system actions as one-hot vectors, and LaRL, HDNO, LAVA treats them as latent variables. Besides, HDSA uses BERT to predict system actions.
DAMD, PARG, SimpleTOD, MarCo, and UBAR treat belief state, system actions as sequences and generate them along with system response. 
Besides, DAMD (aug) means DAMD using rule-based multi-action data augmentation to augment the system actions. Similar to HDSA, MarCo also uses BERT to predict system actions.

\subsection{Overall Results}

As shown in Table \ref{tab:overall_results}, our model significantly outperforms the baseline model DAMD in \textit{Inform Rate}, \textit{Success Rate} and especially \textit{Combined Score}. Besides, our model achieves the best performance in \textit{Combined Score} among all the baseline models. We also observe that models that generate system actions as a sequence generally have superior performance, implying that sequence is a better representation to model the inter-relationships among dialogue actions than one-hot vectors.
Besides, our model outperforms all the methods with data augmentation, which shows the effectiveness of our proposed retrieve-and-memorize framework.

We also evaluate our model on MultiWOZ 2.1 \cite{eric2020multiwoz}, an updated version of MultiWOZ 2.0. As shown in Table \ref{tab:overall_results_2.1}, the results are consistent with that on MultiWOZ 2.0 in Table \ref{tab:overall_results}.

\subsection{Performance Across Different Domains}

\begin{figure}[h]
    \centering
    \includegraphics[scale=0.25]{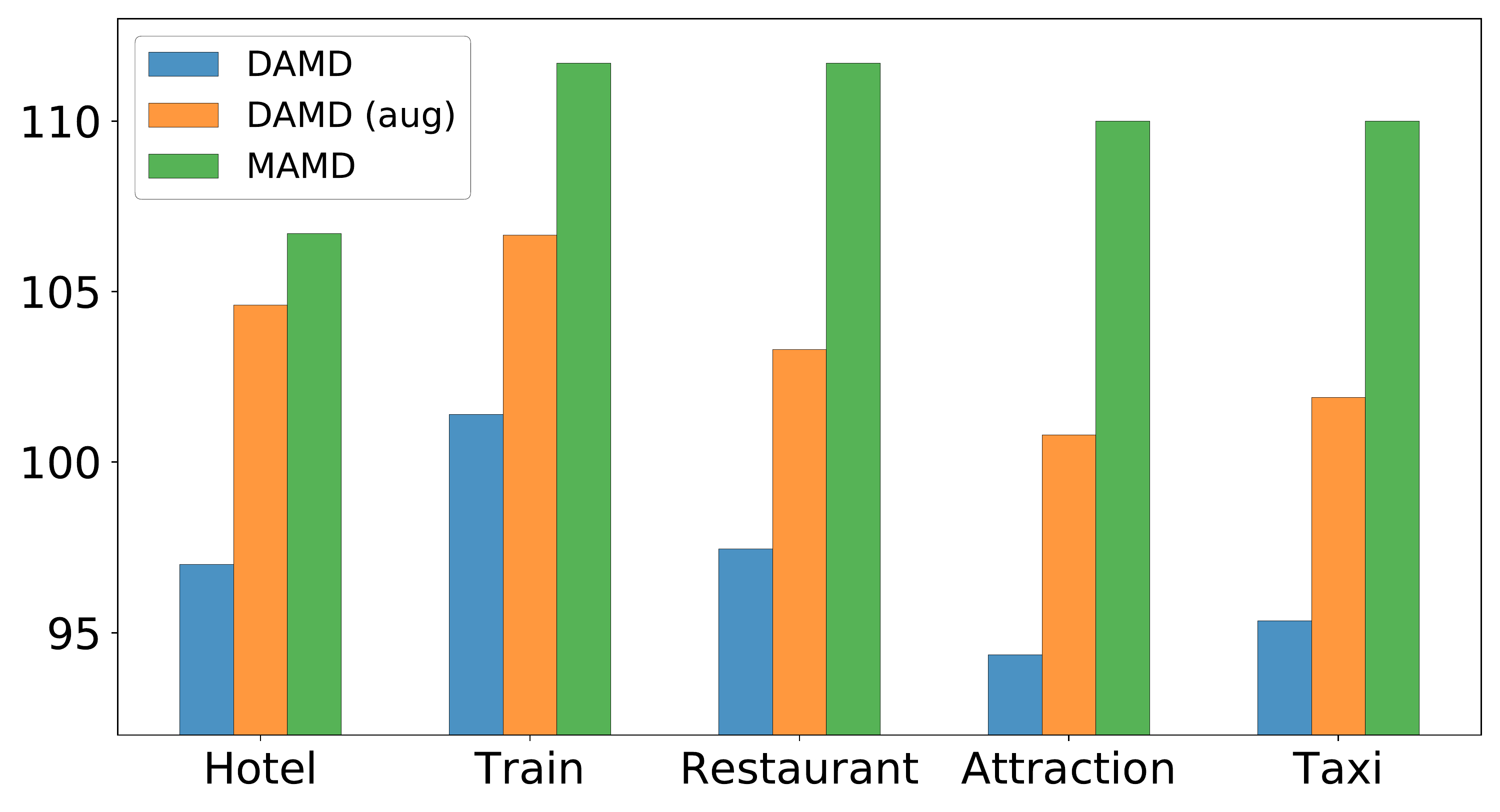}
    \caption{Results of our MAMD and DAMD in combined scores across different domains.~If a dialogue involves more  than one domain, it is counted into each.}
    \label{fig:bar}
\end{figure}

We report the performance of our model on different domains of MultiWOZ 2.0 and compare it with DAMD and DAMD (aug). The results are shown in Figure \ref{fig:bar}. From the bar chart, we can find that our model achieves the best performance across all domains. 
Besides, our model achieves significant performance improvements in \textit{taxi} and \textit{attraction} domains, which appear less frequently in the training data than other domains. Our MAMD narrows the performance gaps among different domains. 

\subsection{Ablation Study}
\label{sec:further analysis}
\begin{table}[h]
\small
\centering
    \begin{tabular}{l c c}
        \toprule
        \textbf{Method} & \textbf{Score} & \bm{$\Delta$} \\
        \midrule
        Baseline & 98.95 & 0\\ 
        \quad \quad + Random & 91.65 & -7.30\\
        \quad \quad + Rule & 102.15 & +3.20\\
        \quad \quad + CARM & 106.65 & +7.70\\
        \quad \quad  + CARM w/o Pt & 102.25 & +3.30\\
        \hline
        \quad \quad  + Random + MA & 98.10 & -0.85\\
        \quad \quad  + Rule + MA & 106.75 & +7.80\\
        
        \quad \quad  + CARM + MA & 108.70 & +9.75\\
        
        \quad \quad  + CARM + MA + RS & 111.20 & +12.25\\
        \bottomrule
    \end{tabular}
    \caption{Results of ablation study. \emph{Baseline} is MAMD without the memory-augmentation component.~\emph{Random} means randomly selected actions, \emph{Rule} is the rule-based augmentation proposed by DAMD, \emph{CARM} is the proposed context-aware retrieval module, and \emph{w/o Pt} means without pre-training before retrieval.~\emph{MA} is the proposed memory-augmentation module, and \emph{RS} is the proposed random sampling technique.}
    \label{tab:ablation}
\end{table}

In this section, we conduct experiments to study the contributions of the proposed context-aware retrieval module and memory-augmented module.

As shown in Table \ref{tab:ablation}, the first group is the baseline directly trained on four types of augmented data, where it treats the augmented actions as equivalent to the golden ones. We observe that the performance drops significantly if the augmented actions are randomly selected, suggesting that the benefit of such data augmentation is strongly subject to the quality of the augmented data.
Additionally, the model trained with CARM outperforms the Rule, which indicates the higher quality of our context-aware retrieved candidates and the effectiveness of the proposed CARM. What's more, removing the system actions prediction pre-training task in CARM causes a performance drop, which demonstrates the necessity to adjust the pre-trained model and obtain more task-related representations.

The second group in Table \ref{tab:ablation} shows the results of the model with the memory-augmented (MA) module trained as well as evaluated with various augmented data. 
First, with MA, our MAMD is much more robust to random noise, only slightly underperforming the baseline. This is because, during training, a model with MA can learn to ignore the noises in the memory and pay less attention to the memory during evaluation.
Second, we see more performance gains with MA from both rule-based and context-aware retrieved candidates, which suggests a model with MA can utilize the candidate system actions more effectively.
Last but not least, with the random sampling mechanism, the performance of our full model further improves.



\subsection{Effect of Random Sampling}

\begin{figure}[t]
    \centering
    \includegraphics[scale=0.43]{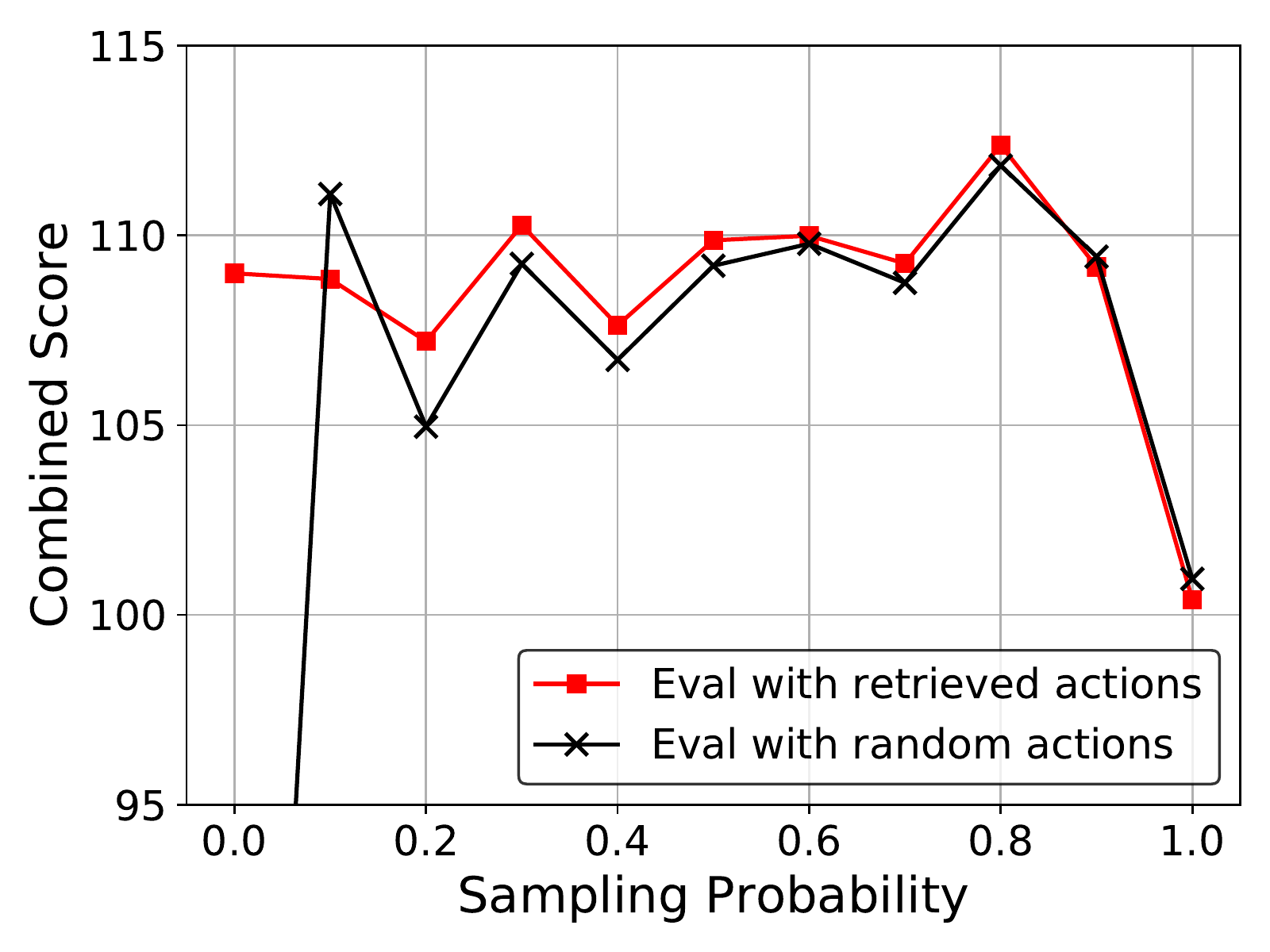}
    \caption{Results of our model trained with different random sampling probabilities and evaluated with different type of candidate actions on the development set.}
    \label{fig:line_shuffle_p}
    \vspace{-0.1cm}
\end{figure}

To further analyze the effect of random sampling, we adjust the random sampling probability during training from 0 (no random sampling and all candidates are from CARM) to 1 (all candidates are randomly sampled), and evaluate MAMD with retrieved candidates and randomly sampled candidates in the memory bank.
As shown in Figure \ref{fig:line_shuffle_p}, the first thing to notice is that without random sampling, i.e., the random sampling probability $p$ is set to 0, the performance of MAMD with random candidate system actions drops drastically to 66.40. This indicates MAMD trained with all decent-quality candidates has developed excessive dependence on the candidates and in a way treats them as ground truth actions, which is what we try to avoid by introducing random sampling. 
Once we introduced random sampling, the performance gap between MAMD evaluated with retrieved actions and random actions is significantly narrowed, which suggests MAMD is capable of telling the quality of the candidates in the memory bank.


\begin{figure}[t]
    \centering
    \includegraphics[scale=0.46]{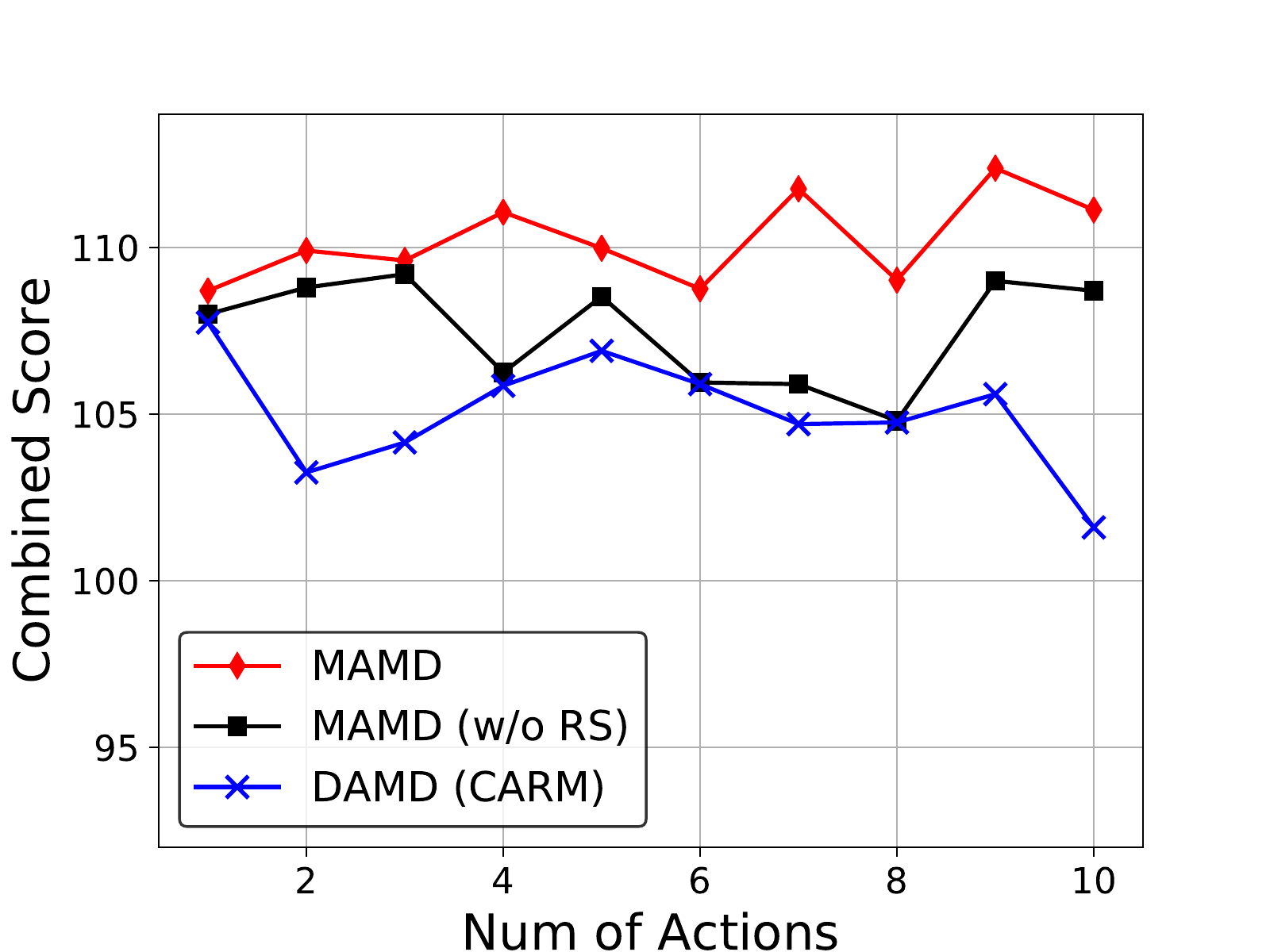}
    \caption{Combined score of three models trained with different numbers of candidate actions retrieved by CARM on the development set, where \emph{MAMD (w/o RS)} means our model without random sampling and \emph{DAMD (CARM)} means DAMD trained with augmented system actions retrieved by CARM.} 
    \label{fig:line_act_score}
\end{figure}

\subsection{Effect of the Number of Candidate Actions}

To analyze the effect of the number of candidate actions on our proposed modules, we train three model variations with different numbers of candidate actions retrieved by CARM. As shown in Figure \ref{fig:line_act_score}, we can see that both MAMD and MAMD (w/o RS) achieve their best performances with 9 candidate actions.
Additionally, both our models consistently outperform DAMD, which suggests the effectiveness of the memory-augmented module.~What's more, the performance of our full model increases more steadily as the number of candidate actions goes up, while without random sampling, the performance of our model is much more unstable across different numbers of candidate actions, which indicates that random sampling can bring in some desirable regularization. 

\begin{figure}[t]
    \centering
    \includegraphics[scale=0.33]{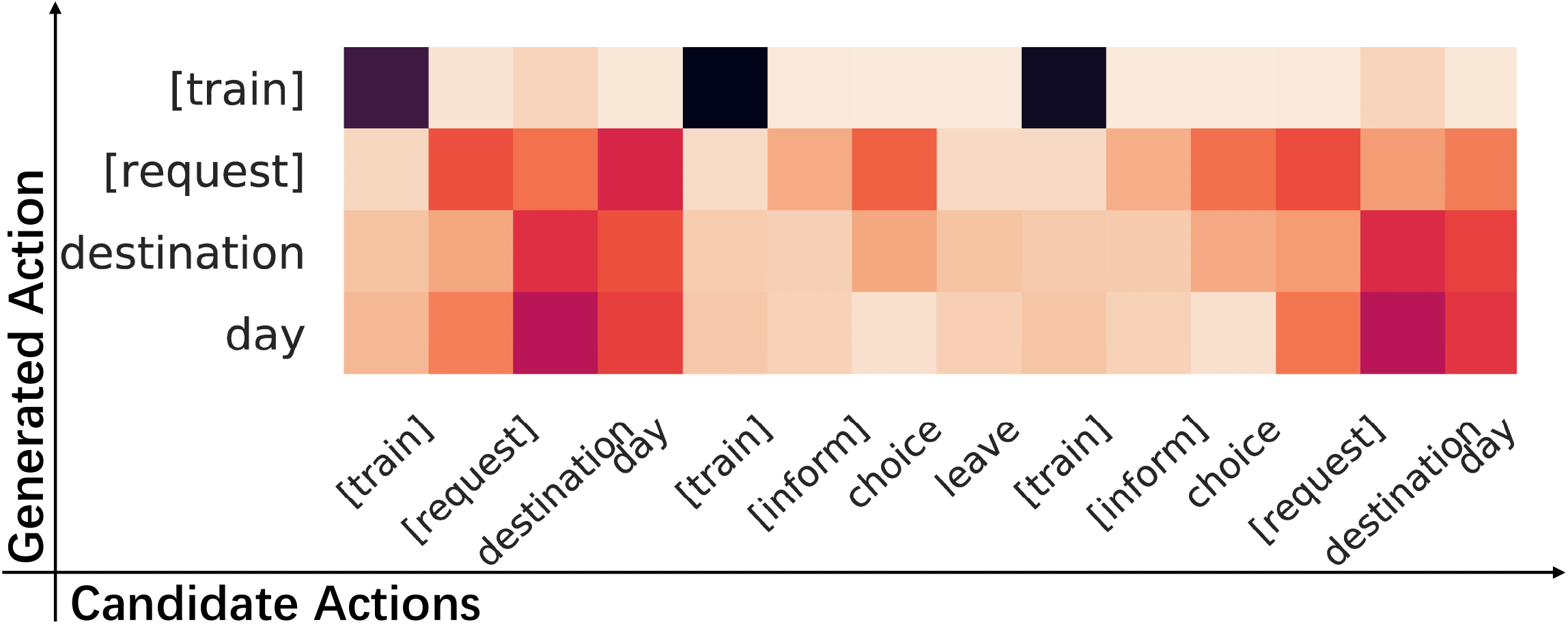}
    \caption{Visualization of the attention from generated system actions to candidate actions. The y-axis is generated system actions and the x-axis is candidate system actions. At each decoding step, the generated system actions selectively attend to the candidate actions. (Dialogue ID:MUL0473)}
    \label{fig:heat_map}
    \vspace{-0.4cm}
\end{figure}
\begin{table}[t]
	\renewcommand{\arraystretch}{1.2}
\footnotesize
    \centering
    \begin{tabular}{p{0.7cm}|p{6cm}}
    \midrule[1.0pt]
    \multicolumn{2}{p{7cm}}{\textbf{Context:} 
    
    \textbf{Sys}: I would recommend the cambridge museum of technology, would you like any information about that? 
    
    \textbf{User}: Yes. What is the postcode and phone number?}\\
    \hline
    \multicolumn{2}{p{7cm}}{\makecell[l]{\textbf{DAMD:}\\$\left[\text{attraction}\right]\left[\text{recommend}\right]$ postcode phone name type\\$\left[\text{attraction}\right]\left[\text{inform}\right]$ phone postcode\\$\left[\text{attraction}\right]\left[\text{nooffer}\right]$ type\\ \textbf{......}}}\\
    \hline
    \multicolumn{2}{p{7cm}}{\makecell[l]{\textbf{CARM:}\\$\left[\text{attraction}\right]\left[\text{inform}\right]$ phone postcode \\    $\left[\text{attraction}\right]\left[\text{inform}\right]$ postcode phone [general] [reqmore]}} \\
    \hline
    \multicolumn{2}{p{7cm}}{\makecell[l]{\textbf{Reference:}\\    $\left[\text{attraction}\right]\left[\text{inform}\right]$ postcode phone [general] [reqmore]}}    \vspace{-0.15cm}\\
    \midrule[1.0pt]
    \end{tabular}
    \caption{Comparison of retrieved candidate system actions of DAMD and our CARM.}
    \label{tab:case}
    \vspace{-0.2cm}
\end{table}
\begin{table}[!ht]
	\renewcommand{\arraystretch}{1.2}
\footnotesize
    \centering
    \resizebox{7.6cm}{2.8cm}{\begin{tabular}{p{0.0cm}|p{6cm}}
    \midrule[1.0pt]
    \multicolumn{2}{p{7cm}}{\textbf{Context:} ... User: Please book tickets and provide me with the total cost of tickets and confirmation number.}\\
    \hline
    \multicolumn{2}{p{7cm}}{\textbf{DAMD:}}\\
    \multicolumn{2}{p{7cm}}{The [value\_id] is \textbf{\underline{[value\_price]}}. The train id is [value\_id]. Is there anything else I can help you with?}\\
    \hline
    \multicolumn{2}{p{7cm}}{\textbf{MAMD:}}\\
    \multicolumn{2}{p{7cm}}{Booking was successful, the total fee is \textbf{\underline{[value\_price]}} payable at the station.~Reference number is: \textbf{\underline{[value\_}} \textbf{\underline{reference]}}.~Is there anything else I can help you with?}\\
    \hline
    \multicolumn{2}{p{7cm}}{\textbf{Reference:}}\\
    \multicolumn{2}{p{7cm}}{It has been booked! Your reference number is \textbf{\underline{[value\_}} \textbf{\underline{reference]}}.~The cost is \textbf{\underline{[value\_price]}}. Do you need anything else?}\\
    \midrule[1.0pt]
    \end{tabular}}
    \caption{An example of response generation of DAMD and  MAMD.}
    \label{tab:case2}
\end{table}

\section{Visualization and Case Study}
\label{sec:case study}
An illustrative example is shown in Figure \ref{fig:heat_map}, the current user utterance is ``\textit{I need a train departing cambridge arriving by 20:30}". The action decoder successfully attends to appropriate actions and ignores the noisy ones like ``\textit{[train] [inform] leave}", as the leaving time has not provided by the user.

Table \ref{tab:case} shows an example of candidate system actions that CARM appropriately retrieved but DAMD failed.~The user asks the system to provide the postcode and phone number of the attraction, while DAMD returns ``\textit{[attraction][nooffer][type]}". 

We also present an example of response generation in Table \ref{tab:case2}, where the user asks for the price and reference number. DAMD manages to provide the postcode but fails to provide the reference number, while our MAMD model successfully provides both the postcode and the reference number. 


\section{Human Evaluation}
\label{sec:human_evaluation}
Finally, we conduct a human study to evaluate our model from the human perspective. We randomly select 30 dialogue sessions (211 dialog turns in total) from the test dataset and have 5 postgraduates as judges to compare two groups of systems: MAMD vs. DAMD and MAMD vs. Reference, in terms of \textit{Readability} and \textit{Completion} \cite{wang-etal-2020-multi-domain}.
\textit{Completion} measures whether a response has correctly answered a user query, including relevance and informativeness.~\textit{Readability} measures the fluency and consistency of the response.

\begin{table}[t]
	\renewcommand{\arraystretch}{1.4}
\small
    \centering
    \begin{tabular}{p{2.9cm}<{\centering} |c c c }
         \midrule[1.0pt]
          \textbf{MAMD vs. DAMD} & \textbf{Win\%} & \textbf{Tie\%} & \textbf{Lose\%} \\
         \hline
         Completion & 19.25\% &66.54\% & 14.21\% \\
         Readability & 3.13\% & 93.08\% & 3.79\% \\
        \midrule[1.0pt]
    \end{tabular}
    \begin{tabular}{c |c c c }
        \midrule[1.0pt]
        \textbf{MAMD vs. Reference} & \textbf{Win\%} & \textbf{Tie\%} & \textbf{Lose\%} \\
        \hline
        Completion & 14.51\% &56.11\% & 29.38\% \\
        Readability & 2.85\% & 92.03\% & 5.11\% \\
        \midrule[1.0pt]
    \end{tabular}
    \caption{Results of human evaluation on response quality. \emph{Reference} means ground truth response. \emph{Win}, \emph{Tie} and \emph{Lose} respectively indicate the proportions that our model wins over, ties with or loses to its counterpart.}
    \label{tab:human}
\end{table}

We report the human evaluation results in Table \ref{tab:human}, from which we can observe that our model outperforms DAMD and beats or ties with Reference nearly 70\% of the time in terms of \textit{Completion}.
In \textit{Readability}, our model ties more than 92\% with DAMD as well as Reference.~This may suggest the language of responses lacks diversity and is easy to learn.
Overall, our model is superior to DAMD in human evaluation, which demonstrates its competence in a more holistic evaluation other than automatic metrics.

\section{Conclusion}
\label{sec:conclusion}
In this paper, we proposed a retrieve-and-memorize framework to deal with the unbalanced distribution of system actions in task-oriented dialogue systems. Our framework includes a neural retrieval module that can retrieve multiple candidate system actions given a dialogue context, and a memory-augmented multi-decoder network that can generate system actions conditioned on multiple candidate system actions. Extensive experiments were conducted on a large-scale multi-domain task dialogue dataset and the results demonstrate the effectiveness of our framework. In essence, the whole framework, including its random sampling strategy, can be viewed as an attempt to prevent the systems from overfitting skewed dialogue datasets with an unbalanced distribution of system actions. 
\section*{Acknowledgments}
The paper was supported by the National Natural Science Foundation of China (No.61906217) and the Program for Guangdong Introducing Innovative and Entrepreneurial Teams (No.2017ZT07X355).
\bibliography{references}
\bibliographystyle{acl_natbib}

\appendix

\section{More Details of MAMD}
\subsection{Attention Function}
In MAMD, we use an attention function \emph{Attn} to attend to three groups of hidden states. In this study, $Attn(h,H_a,H_b,H_c)$ is defined as:
\begin{equation}
    \begin{aligned}
    &h_a = \text{CatAttn}(h,H_a),\\
    &h_b = \text{CatAttn}(h,H_b),\\
    &h_c = \text{CatAttn}(h,H_c),\\
    &\text{Attn}(h,H_a,H_b,H_c) = [h_a \oplus h_b \oplus h_c],
    \end{aligned}
\end{equation}
where $\oplus$ is the concatenation operator and CatAttn is a simple concat-attention defined as:
\begin{equation}
    \begin{aligned}
        &a_i            = \text{tanh}(W[h \oplus H_i]),\\
        &{\alpha}_{i}   = \text{Softmax}(a),\\
        &\text{CatAttn}(h,H) = \sum\nolimits_{i=1}^{n}{{\alpha}_{i} H_i},\\
    \end{aligned}
\end{equation}
where W represents learnable parameters, $H$ is the sequence of encoded hidden states,\footnote{For example, the encoded hidden states of user utterance.} and $n$ is the number of hidden states in $H$.

\subsection{Decoder with Copy Mechanism}
The decoder used to generate the belief state, system action and response is a one-layer GRU augmented with copy mechanism. Each step of the generation in Dec$(c_t,h_{t-1},H)$ is defined as follows:
\begin{equation}
    \begin{aligned}
        &h_t        = \text{GRU}(c_t,h_{t-1}),\\
        &p_{vocab}  = \text{Softmax}(W_vh_t),\\
        &s_i        = h_t^{\top}\text{tanh}(W_cHi),\\
        &p_{copy}   = \text{Softmax}(s),\\
        &p_{final}(w)       = p_{vocab}(w) + \sum_{i:X(i)=w}{p_{copy}^i},\\
        &\text{Dec}(c_t,h_{t-1},H) = p_{final},h_t,
    \end{aligned}
\end{equation}
where $W_v$ and $W_c$ are learnable weights, and X is the corresponding context of H.

\section{More implementation Details}
\subsection{Hyperparameters}
In this section, we report the hyperparameter setting in our model. For MAMD, we adopt the default hyperparameters in DAMD, as shown in Table \ref{tab:hyper}. As for the learning rate, the number of candidate actions, and the random sampling probability, we apply grid search to find the best combination on the development set. It takes about 10 hours to train our model on a single  12 GB Nvidia GeForce RTX 2080 Ti. As for CARM's pre-training task, the hyperparameter setting is shown in Table \ref{tab:hyper_CARM}.
\begin{table}[h]
\small
    \centering
    \begin{tabular}{l |c}
    \toprule
         \textbf{Parameter}& \textbf{Values} \\
    \hline
         batch size                     & 80 \\
         learning rate                  & 7e-3 \\
         embedding size                 & 50 \\
         hidden size                    & 100 \\
         dimension of db search result  & 6 \\
         encoder layers                 & 1\\
         decoder layers                 & 1\\
         epoch                          & 60 \\
         candidate actions              & 9 \\
         random sampling probability    & 0.8 \\
         beam size                      & 5 \\
         random seed                    & 777 \\
    \bottomrule
    \end{tabular}
    \caption{Hyperparameter setting of MAMD.}
    \label{tab:hyper}
\end{table}
\begin{table}[h]
\small
    \centering
    \begin{tabular}{l |c}
    \toprule
         \textbf{Parameter}& \textbf{Values} \\
    \hline
         batch size                     & 6 \\
         learning rate                  & 5e-5 \\
         epoch                          & 20 \\
         random seed                    & 42 \\
         max sequence length            & 400 \\
         warmup proportion              & 0.1 \\
    \bottomrule
    \end{tabular}
    \caption{Hyperparameter setting of CARM.}
    \label{tab:hyper_CARM}
\end{table}

\subsection{Delexicalization Strategy}
For delexicalization, we follow DAMD's domain-adaptive delexicalization strategy. Specially, we use tokens such as [value\_name] to represent the same slot name. In this case, the placeholders [hotel\_name] and [restaurant\_name] will be converted to [value\_name]. During the evaluation, we induce the domain of a placeholder from the transition between two adjacent belief states and the generated system actions of the current dialog turn.\footnote{For more details, please refer to the source code.}

\subsection{Post-Processing of Candidate Action Retrieval}
As for candidate action retrieval, we retrieve 50 candidate actions for each sample. Then, we apply post-processing to clean the candidate actions: 
\begin{itemize}
\item Duplicated actions are merged. For example, the system actions ``[attraction] [inform] postcode phone [general] [reqmore]'' and ``[attraction] [inform] postcode phone [general] [reqmore]'' will be combined into ``[attraction] [inform] postcode phone [general] [reqmore]''.
\item Null system actions are removed.
\item System actions with different database query results are filtered out.
\item System actions that conflict with current belief  are filtered, e.g., requesting a slot that is already included in belief states. 
\end{itemize}

\section{Dataset Details}
\begin{figure}[h]
    \centering
    \includegraphics[scale=0.60]{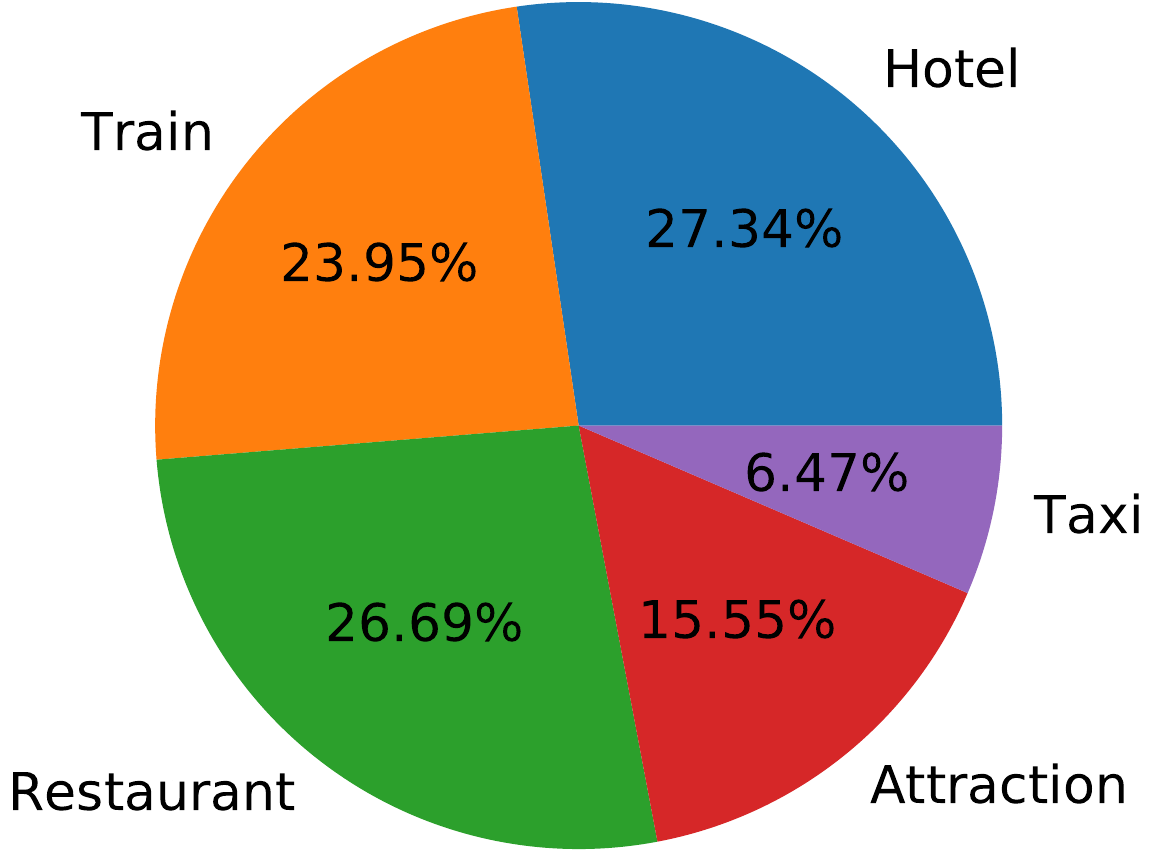}
    \caption{Statistics of system actions across different domains of MultiWOZ 2.0.} 
    \label{fig:domain}
\end{figure}

We provide more information about the MultiWOZ 2.0 dataset. 
The training set contains 8438 dialogs, 115,424 turns, and  1,520,970 tokens. The average number of turns per dialog is 13.68, and the average number of tokens per turn is 13.18. The number of slots and values are 25 and 4510, respectively. The ontology is shown in Table \ref{tab:ontology}. We also count the numbers of system actions across different domains. As shown in Figure \ref{fig:domain}, the numbers of system actions in \textit{attraction} and \textit{taxi} are smaller than the other domains, showing the unbalanced distribution of system actions at the domain level.  

\begin{table}[h]
    \centering
    \resizebox{7.5cm}{2.5cm}{
    \begin{tabular}{|c|c|}
        \hline
        \multirow{3}{1.2em}{act type}  & inform$^*$ / request$^*$ / nooffer$^{1234}$ / \\
        & recommend$^{123}$ / select$^{1234}$ / offerbook$^{124}$ /  \\
        & offerbooked$^{124}$ / nobook$^{12}$ / bye$^*$ / greet$^*$ / \\ 
        & reqmore$^*$ / welcome$^*$ \\ 
        \hline
        \multirow{3}{1.2em}{slot}  & car$^5$ / address$^{12367}$ / postcode$^{12367}$ / \\ 
        & phone$^{123567}$ / internet$^2$ / parking$^2$ / type$^{23}$ / \\ 
        & pricerange$^{12}$ / food$^1$ / stars$^2$ /  area$^{123}$ / \\
        & reference$^{1234}$ / time$^{14}$ / \\
        & leave$^{45}$ / price$^{45}$ / arrive$^{45}$ / id$^4$ / \\ 
        & stay$^2$ / day$^{124}$ / leave$^{45}$ / people$^{123}$ / name$^{123}$ / \\
        & destination$^{45}$ / departure$^{45}$ / department$^6$ \\
        \hline
    \end{tabular}}
    \caption{Ontology for all domains. The upper script indicates which domains it belongs to ($*$: universal, 1: restaurant, 2: hotel, 3: attraction, 4: train, 5: taxi, 6: hospital, 7: policy).}
    \label{tab:ontology}
\end{table}

\begin{table}[!h]
\small
\centering
\setlength{\tabcolsep}{2.0mm}{
    \begin{tabular}{l c c c c}
    \toprule
        \textbf{Dateset} & \textbf{Inform} & \textbf{Success} & \textbf{BLEU} & \textbf{Score} \\
    \midrule
        \textbf{Development}    & 96.60 & 90.70 & 18.70 & 112.35 \\
        \textbf{Test}           & 95.70 & 88.90 & 18.90 & 111.20 \\
    \bottomrule
    \end{tabular}
}
    \caption{Overall results on the MultiWOZ 2.0 dataset.}
    \label{tab:overall_results_2.0}
\end{table}

\begin{table}[!h]
\small
\centering
\setlength{\tabcolsep}{2.0mm}{
    \begin{tabular}{l c c c c}
    \toprule
        \textbf{Dateset} & \textbf{Inform} & \textbf{Success} & \textbf{BLEU} & \textbf{Score} \\
    \midrule
        \textbf{Development}    & 94.90 & 87.70 & 18.60 & 109.90 \\
                \textbf{Test}           & 94.20 & 86.20 & 18.80 & 109.00 \\
    \bottomrule
    \end{tabular}
}
    \caption{Overall results on the MultiWOZ 2.1 dataset.}
    \label{tab:overall_results_2.1}
\end{table}

\section{More Analyses and Discussions}

\subsection{Results on Development and Test Sets}
We report the results of MAMD on the development and test sets of MultiWOZ 2.0 and MultiWOZ 2.1. As shown in  Table \ref{tab:overall_results_2.0} and Table \ref{tab:overall_results_2.1}, the results on the development set are generally consistent with that on the test set on both benchmarks.



\subsection{Distribution of Generated System Actions}
To further analyze the influence of our model on the generation of system actions, we count the appearance of generated actions. Recall that each dimension of the actions stands for either \textit{domain}, \textit{function} or \textit{slot}, where \textit{domain} defines the domain involved in the conversation, and \textit{function} defines the behavior of system such as informing the user or request certain information. Here we only count the first two dimensions of the actions because the third dimension appears to be less important.

\begin{figure*}[h]
    \centering
    \includegraphics[scale=0.55]{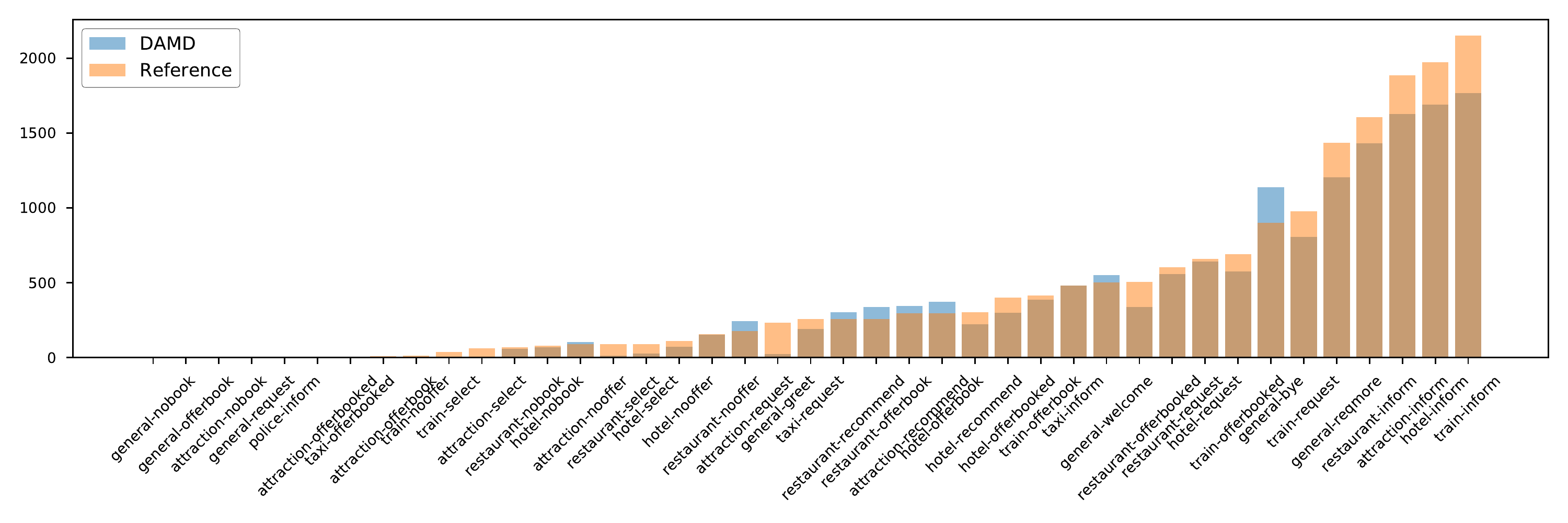}
    \includegraphics[scale=0.55]{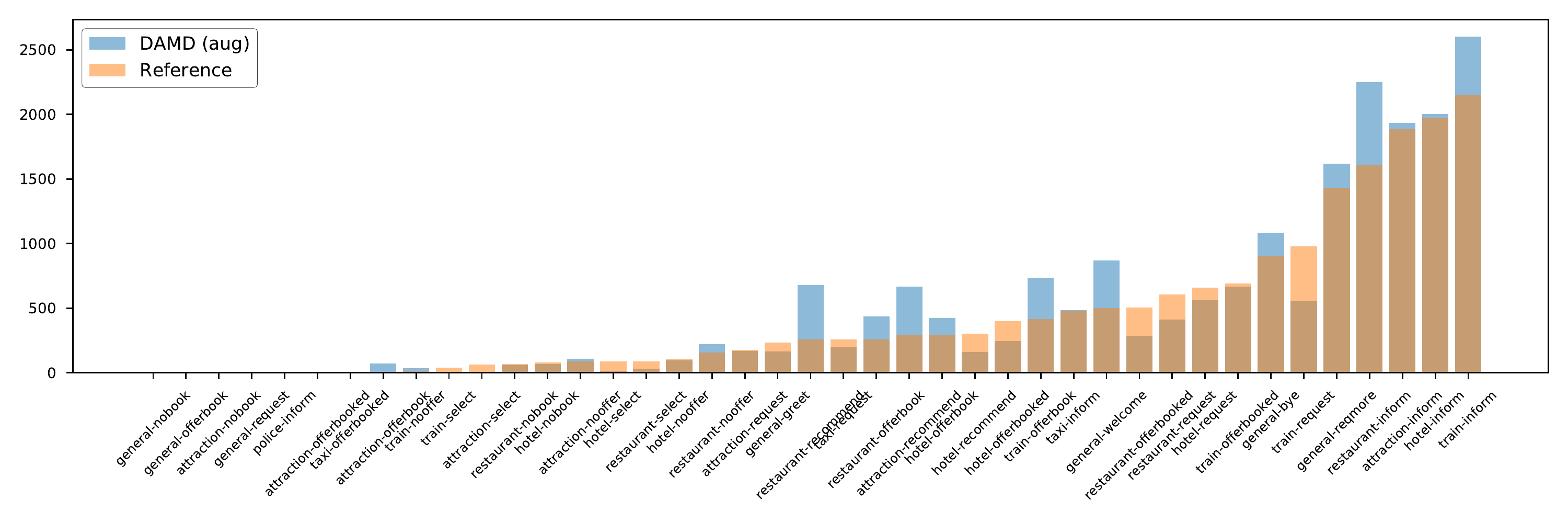}
    \includegraphics[scale=0.55]{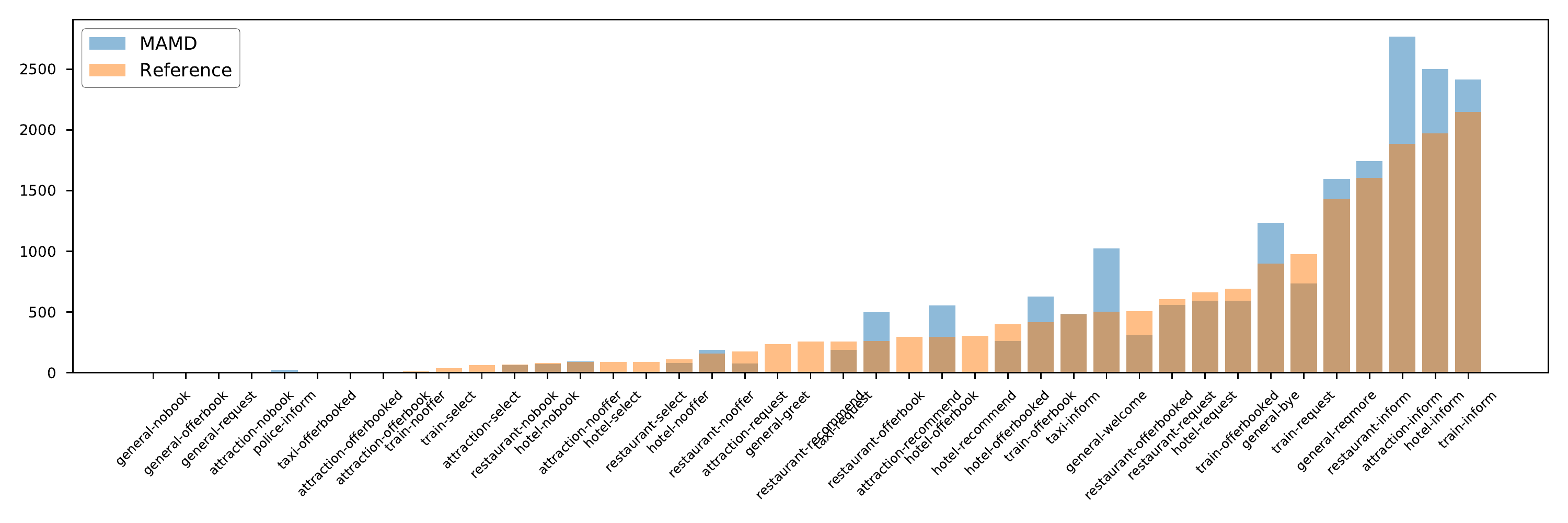}
    \caption{Statistics of generated system actions by DAMD, DAMD (aug) and MAMD, and comparison with reference actions.} 
    \label{fig:act_dist}
\end{figure*}

As shown in Figure \ref{fig:act_dist}, the distribution of system actions generated by DAMD is proportional to the original distribution in the dataset, and DAMD tends to generate fewer actions than the original distribution.
After applying their rule-based multi-action data augmentation, DAMD (aug) can generate more diverse system actions compared with DAMD. Compared with DAMD (aug), MAMD generates more actions. More importantly, MAMD generates more important actions such as ``attraction-inform'' and ``taxi-inform'' which are more relevant to task completion, while DAMD (aug) tends to generate less useful actions such as ``general-require'' and ``general-greet''.
This phenomenon indicates that the memory-augmented mechanism provides some guidance to our model during system action learning. 
To sum up, our proposed model can generate more diverse and valuable actions, which demonstrates the effectiveness of our proposed memory-augmented mechanism.

\end{document}